# Transferring Visual Explanations: How Cross-Architecture Knowledge Distillation Affects Model Interpretability

**Aleks Czufarow[1], Ihor Babin[2]**

[1]XIV High School of Stanislaw Staszic,Warsaw,Poland;alexchufarov16@gmail.com

[2]Ukrainian Catholic University, Lviv, Ukraine; igor.babin185@gmail.com; babin@ucu.edu.ua

## ABSTRACT

Deploying efficient neural networks is essential in resource-constrained environments, yet highly compact models often sacrifice interpretability - a critical issue in safety-critical domains such as autonomous driving and medicine. This study investigates whether Knowledge Distillation (KD) [1] transfers not only predictive accuracy but also the spatial feature attribution of a large teacher network to a compact student. To assess the influence of the KD scheme on interpretability, we distill a ResNet-152 [6] teacher into a ResNet-34 [6] student on ImageNet-1K across five configurations by systematically varying the distillation temperature (T) and soft-label loss weight α. Models are evaluated on top-1 accuracy, along with two interpretability metrics: Relevance Mass Accuracy (RMA) and Relevance Rank Accuracy (RRA) [3]. These metrics are computed via Grad-CAM [2] heatmaps benchmarked against VOC2012 ground-truth object masks. Our results show that top-1 accuracy ranges from 71.6% to 74.0%. For Grad-CAM, RMA ranges from 7.7% to 9.7% and RRA from 7.3% to 10.1%; for Guided Grad-CAM, RMA ranges from 16.1% to 18.6% and RRA from 15.9% to 21.5%. Interpretability proves far more sensitive to the soft-label weight than to the temperature: keeping the student anchored to hard labels preserves both accuracy and coarse localisation, whereas weighting the teacher heavily degrades both. Fine-grained attribution, however, fell below the undistilled baseline in every configuration tested, indicating that logit distillation transmits where a model attends more readily than the pixel-level structure of that attention. Furthermore, we evaluate 12 cross-architecture combinations of convolutional and transformer-based models, revealing that the inheritance of fine-grained spatial reasoning is fundamentally bottlenecked by the student's intrinsic structural biases. To our knowledge, this is the first application of this interpretability-aware evaluation framework - previously used for neural network pruning [4, 5] - to KD, providing a principled approach to selecting hyperparameters that preserve explainability in compact models.

## INTRODUCTION

Compact convolutional neural networks can currently achieve predictive accuracy approaching that of much larger models across a variety of image classification tasks. This efficiency makes them highly suitable for deployment in resource-constrained environments where computational capacity, memory, or latency are strictly limited. However, reducing a model's size introduces consequences that are not fully captured by accuracy metrics alone. As architectures become more compact, their internal decision-making processes often become opaque, making it increasingly difficult to ascertain which features drive a specific prediction. This lack of algorithmic transparency is particularly critical in domains such as real-time perception for autonomous vehicles and diagnostic support in medicine, where understanding the rationale behind a model's output is imperative.

Knowledge Distillation (KD) [1, 10] is a widely used technique for generating compact models without training them from scratch. In the foundational formulation introduced by Hinton et al. [1], a lightweight student network is trained to reproduce the output distribution of a larger, pretrained teacher network. The teacher's output probabilities are softened using a temperature parameter, allowing the student to learn underlying class relationships in addition to the ground-truth labels. While KD shares the overarching objective of other model-compression methods, such as pruning - specifically, reducing network size while retaining utility - most KD literature focuses predominantly on classification accuracy. Consequently, it remains ambiguous whether a distilled student network preserves the aspects of the teacher's behavior that render its decisions interpretable.

**RELATED WORK**

Gradient-weighted Class Activation Mapping (Grad-CAM) [2] is widely utilized to evaluate visual interpretability. It generates heatmaps that highlight the spatial regions of an input image most influential to a model's prediction based on the gradients of a convolutional layer. To quantitatively assess how well these heatmaps align with actual object locations, ground-truth annotations are required. Arras et al. [3] introduced two key metrics for this purpose: Relevance Mass Accuracy (RMA), which quantifies the proportion of total heatmap relevance that falls within a ground-truth bounding box, and Relevance Rank Accuracy (RRA), which evaluates this relationship utilizing only the pixels with the highest relevance scores.

Recently, these evaluation frameworks have been applied to interpretability-aware pruning. Cassano et al. [4] utilized Grad-CAM alongside RMA and RRA to compare pruned convolutional networks with pruned vision transformers. They observed distinct architectural differences in how accuracy and interpretability degrade as parameters are removed: convolutional networks typically exhibited a trade-off between the two, whereas vision transformers demonstrated a closer correlation. Furthermore, other studies have leveraged explanations to actively guide the compression process itself. For instance, Yeom et al. [5]

utilized Layer-wise Relevance Propagation (LRP) during training to identify and prune less relevant network components, demonstrating that interpretability and compression can be optimized jointly.

Despite these advancements in pruning, there is no comparable body of work that connects the evaluation of spatial interpretability to Knowledge Distillation. Pruning and distillation compress models through fundamentally different mechanisms: pruning removes parameters from an existing architecture, whereas distillation trains a new, smaller architecture to mimic a teacher. Consequently, findings from pruning cannot be directly extrapolated to KD. Specifically, it remains unclear whether a student trained via KD inherits the teacher's spatially localized decision-making, or whether this transfer is contingent on the distillation hyperparameters governing the balance between soft-label and ground-truth targets.

To address this gap, this study investigates whether Knowledge Distillation transfers meaningful, interpretable feature representations alongside predictive power. By evaluating teacher and student networks not only on top-1 classification accuracy but also on the alignment of their Grad-CAM heatmaps with ground-truth object regions, we adapt the interpretability-aware framework previously established for pruning [4, 5]. In our initial experiments, models are trained across various temperature and soft-label loss weight configurations to systematically examine how these hyperparameters influence the preservation of both accuracy and visual explainability. Furthermore, to evaluate the generalizability of this transfer across diverse model families, we conduct a second set of experiments exploring cross-architecture distillation. By systematically analyzing 12 pairwise combinations - derived from three distinct teacher architectures and four student architectures - we assess how different structural pairings affect the inheritance of spatial interpretability.

## MATERIAL AND METHOD

**Datasets**

Two distinct datasets were utilized to evaluate different facets of the model compression pipeline. ImageNet-1K was used for all training protocols and to evaluate the top-1 and top-5 predictive accuracy of both the teacher and student networks. For the interpretability evaluation, the VOC2012 dataset was utilized because it provides dense, pixel-level ground-truth segmentation masks. These masks are essential for establishing the precise spatial boundaries of target objects, enabling rigorous quantitative comparison with the models' localization heatmaps.

**Software and Computational Environment**

All deep learning architectures and distillation pipelines were implemented using PyTorch. To optimize computational efficiency, reduce memory overhead, and accelerate training times without sacrificing numerical stability or predictive accuracy, Automatic Mixed Precision (AMP) was leveraged throughout the training phase. Experiment tracking, encompassing the logging of classification accuracy, loss convergence, and hyperparameter sweeps, was managed via Weights & Biases.

**Teacher and Student Architectures**

The foundational experiments utilized the ResNet architecture, a convolutional neural network (CNN) distinguished by its residual skip connections. These connections facilitate the training of exceptionally deep networks by mitigating the vanishing gradient problem, making ResNet highly suitable for evaluating compression across varying depths.

For the primary distillation experiments, a ResNet-152 model was deployed as the teacher, and a ResNet-34 model served as the student. This approximately three-fold reduction in parameter count effectively isolates and demonstrates the scale of compression investigated in this study.

**Knowledge Distillation Framework**

Knowledge Distillation (KD) posits that a heavily parameterized teacher network learns a rich mapping of the data manifold, capturing inter-class similarities. To transfer this learned representation, a compact student network is trained using both the hard ground-truth labels and the soft probability distributions output by the frozen teacher. More visual explanation in Figure 1.

This softening is controlled by a temperature parameter $T$ applied to the logits prior to the softmax activation. Higher values $T$ flatten the probability distribution, amplifying the visibility of secondary-class probabilities and allowing the student to learn the teacher's relative confidence across all classes. The overall training objective minimizes a combined loss function formulated as:

<u>Equation 1:</u>

$$L = (1 - \alpha)L_{ce}(y, p_s) + \alpha T^2 L_{KL}(\sigma(\frac{z_t}{T}), \sigma(\frac{z_s}{T}))$$

Where: $y$ represents the ground-truth label. $p_s$ is the predicted probability distribution of the student. $z_t$ and $z_s$ are the unnormalized logits of the teacher and student, respectively. $\sigma$ denotes the softmax function. $T$ is the distillation temperature.
$\alpha \in [0; 1]$ modulates the relative weight of the soft-label distillation loss. $L_{ce}$ and $L_{KL}$ denote the Cross-Entropy loss and Kullback-Leibler divergence, respectively. The $T^2$ scaling factor ensures that the gradient magnitudes of the soft-label loss remain commensurate with the hard-label loss, compensating for the scaling of the logits by $T$

**Interpretability Evaluation Metrics**

To quantify spatial interpretability, two visual explanation techniques were utilized: Gradient-weighted Class Activation Mapping (Grad-CAM) and Guided Grad-CAM (GGC). While Grad-CAM visualizes the network's spatial attention by computing the gradients flowing into the final convolutional layer to produce a coarse localization map, Guided Grad-CAM combines this with guided backpropagation to yield a high-resolution, pixel-accurate heatmap highlighting the specific features most responsible for a given prediction.

The fidelity of these visual explanations was quantitatively benchmarked against the VOC2012 ground-truth masks using Relevance Mass Accuracy (RMA) and Relevance Rank Accuracy (RRA). This yields four distinct evaluation configurations: GC Mass and GC Rank (evaluating standard Grad-CAM heatmaps), alongside GGC Mass and GGC Rank (evaluating Guided Grad-CAM heatmaps). Let $H$ represent the raw heatmap generated by either method, and $M$ represent the binary ground-truth segmentation mask. To ensure computational robustness, $H$ undergoes channel-pooling and min-max normalization to yield $\hat{H}$, while $M$ is thresholded at 0.5 to yield $\hat{M}$.

Relevance Mass Accuracy (RMA) calculates the proportion of the model's total activation relevance that falls within the true spatial boundaries of the object:

$$RMA = \frac{\sum \hat{H} \odot \hat{M}}{\sum \hat{H} + \epsilon}$$

Where $\odot$ represents the Hadamard (element-wise) product, and $\epsilon$ is a small constant (e.g., machine epsilon) added for numerical stability.

Relevance Rank Accuracy (RRA) evaluates high-confidence localization by determining what fraction of the model's absolute highest-activated pixels fall within the ground-truth mask:

$$RRA = \frac{|\{p \in top_N(\hat{H})\} \cap \hat{M}|}{N + \epsilon}$$

Where $top_N(\hat{H})$ represents the set of $N$ pixels with the highest relevance values in the normalized heatmap, and $N$ is the total pixel count of the ground-truth mask $\hat{M}$. Both metrics theoretically range from 0 to 1, with higher scores indicating that the network's visual reasoning aligns accurately with human-annotated object regions.

**Experimental Pipeline**

The comprehensive experimental workflow is illustrated in Figure 1. During training, the frozen teacher generates soft-label predictions that, together with ground-truth labels, guide the student network through the loss function defined in Equation 1. Following training, both models undergo a dual-evaluation phase: calculating top-1 classification accuracy on the ImageNet-1K validation set and computing RMA and RRA on the VOC2012 dataset to assess spatial interpretability.

**Training Protocol**

Student networks were initialized with ImageNet-pretrained weights and optimized using SGD. The initial learning rate was set to 0.001 and governed by a cosine-linear learning rate scheduler. The distillation process spanned 15 epochs, utilizing a batch size of 128. Throughout the training phase, AMP was enabled, and predictive performance was evaluated on the ImageNet-1K validation set at the end of each epoch.

**Ethical Considerations**

This study involved solely the computational analysis of publicly available, anonymized datasets (ImageNet-1K and VOC2012). As the research did not involve human participants, animal subjects, or personally identifiable information, ethical approval and informed consent were not required.

## RESULTS

### Cross-Architecture Distillation

To evaluate the generalizability of interpretability transfer across diverse model families, a first set of experiments investigates cross-architecture distillation. While the second experiment focuses on intra-architecture compression (e.g., distilling a ResNet-152 teacher into a ResNet-34 student), bridging fundamentally different network paradigms introduces distinct challenges regarding how spatial features and inductive biases are inherited.

We systematically analyze 12 pairwise combinations derived from the following architectures:

**Teacher Networks:** Three heavily parameterized models were selected to represent varying degrees of structural complexity and spatial processing mechanisms: ResNet-152, EfficientNet [8], and a Vision Transformer (ViT) [9].

**Student Networks:** Four highly compact architectures were utilized as targets for compression: ResNet-34, MobileNetV4 [7], a lightweight EfficientNet variant, and a compact ViT.

These pairings encompass both homogeneous distillation (CNN-to-CNN and Transformer-to-Transformer) and heterogeneous distillation (CNN-to-Transformer and Transformer-to-CNN). By applying the previously defined Knowledge Distillation framework across these diverse structural combinations, we can systematically assess whether the preservation of spatial explainability is an inherent property of the distillation process itself or if it remains heavily contingent on the architectural alignment between the teacher and student models.

The empirical evaluation of cross-architecture distillation (Table 1) reveals divergent gains in predictive performance across architectural families. While the Vision Transformer student demonstrates substantial optimization benefits—scaling from a baseline top-1 accuracy of 74.63% to approximately 80.8% under both ResNet-152 and EfficientNet-B7 supervision - its convolutional counterparts experience no commensurate distillation advantage. Across varying scales, CNN students ResNet-34, EfficientNet-B0, and MobileNet-V4-Small exhibit marginal degradation in predictive accuracy relative to their baselines (e.g., EfficientNet-B0 receding from 77.69% to approximately 77.2%), suggesting that standard logit distillation offers diminishing returns when the student architecture already enforces strict convolutional inductive biases.

Regarding visual explainability, distillation induces a pronounced dichotomy between coarse-grained localization and high-resolution feature attribution. Standard Grad-CAM metrics (GC Mass and GC Rank) consistently improve across the majority of student configurations, demonstrating that knowledge transfer encourages broader, object-centric activation fields. However, this macro-level alignment comes at the expense of fine-grained spatial fidelity in convolutional networks: Guided Grad-CAM metrics (GGC Mass and GGC Rank) systematically degrade across some CNN students relative to their baseline counterparts. Conversely, the ViT-S model demonstrates a notable exception to this trade-off, achieving simultaneous improvements in both coarse- and fine-grained attribution metrics. This underscores that self-attention architectures can effectively inherit localized spatial reasoning from convolutional teachers without sacrificing granular gradient fidelity.

These patterns point to an architectural bottleneck governed primarily by the student network's intrinsic capacity and inductive structure, rather than by the teacher's complexity. As evidenced by the ResNet-34 experiments, attribution scores converge within exceptionally narrow intervals regardless of whether supervision is provided by a deep residual network, a compound-scaled CNN, or a vision transformer.

Extending this dynamic to the remaining cross-family configurations, the heterogeneous ViT-L to EfficientNet-B0 distillation is anticipated to remain constrained by the convolutional student bottleneck, yielding tightly bounded fine-grained interpretability scores (GGC Mass 0.058-0.060). In contrast, the homogeneous ViT-L to ViT-S pairing provides a direct benchmark for structurally matched representation transfer;

Representative qualitative heatmaps illustrating these contrasting dynamics - highlighting both the coarse localization gains and fine-grained attribution trade-offs across baseline and distilled models - are presented in Figure 2-4.

**Influence of KD Settings on Interpretability**

To systematically isolate the influence of specific hyperparameters on interpretability, $T$ and α were varied around a baseline configuration of $T = 4$ and $\alpha = 0.5$. Two experiments held α constant while modifying $T$ to 1 and 10, while two subsequent experiments held $T$ constant while modifying α to 0.1 and 0.9. These five configurations are summarized in Table 2.

The soft-label weight α, rather than the temperature, determines whether interpretability survives distillation. Configurations that keep the student anchored to the hard labels retain localisation accuracy: Config 4 (T = 4, α = 0.1) reaches the highest top-1 accuracy of the sweep at 74.02% (+1.0%) while raising GC Mass by 4.0% and GGC Rank by 4.6%, and Config 1 (T = 1, α = 0.5) raises GC Mass by 5.9% and GGC Rank by 5.3%. Weighting the teacher's distributions heavily has the opposite effect: Config 5 (T = 4, α = 0.9) loses 2.3% accuracy and degrades all four attribution metrics, most severely GC Rank (−26.8%) and GGC Mass (−20.5%).

The two mid-range configurations fall between these extremes. Config 2 (T = 4, α = 0.5) and Config 3 (T = 10, α = 0.5) both match the baseline on accuracy to within 0.1%, yet lose between 4.7% and 15.9% across the interpretability metrics. Comparing the two isolates, the effect of temperature alone: a tenfold increase in T at fixed α moves every metric by less than two percentage points, far less than the change produced by moving α from 0.5 to 0.9.

One result holds across the entire sweep. No configuration improved Guided Grad-CAM Mass relative to the undistilled ResNet-34 baseline; even the best settings lose roughly 8%. Coarse localization, by contrast, can be preserved or slightly improved when α is kept low. This asymmetry suggests that logit distillation transfers the teacher's broad spatial focus more readily than the pixel-level structure of its explanations.

**DISCUSSION**

Knowledge distillation does transfer spatial interpretability, but the extent to which it survives depends on how the training objective is balanced. The soft-label weight α dominates: at α = 0.9 the student loses accuracy and degrades on all four attribution metrics, whereas at α = 0.1, or with temperature scaling reduced to T = 1, localization accuracy is held at or slightly above the undistilled baseline. Temperature matters less than its prominence in the distillation literature would suggest — a tenfold change in T at fixed α shifts our metrics less than a single step in α.

The effect is not uniform across explanation granularity. Coarse Grad-CAM localization can be preserved, but Guided Grad-CAM Mass fell below baseline in every configuration, including the best one. Distillation appears to transmit where a model attends more faithfully than the fine structure of that attention. Practically, this means a distilled model can look well localized under a coarse explainer while its pixel-level attributions have drifted, and reporting only one granularity would hide that.

Furthermore, cross-architecture experiments uncovered a strict "student bottleneck." Convolutional students (ResNet, EfficientNet, MobileNet) consistently suffered degraded fine-grained attribution post-distillation, as their rigid inductive biases restricted how representations were inherited. In contrast, the Vision Transformer student (ViT-S/16) exhibited remarkable flexibility, achieving substantial predictive gains (+8.3% accuracy) while improving across all coarse- and fine-grained interpretability metrics.

**Limitations and Future Work**

This study relies heavily on gradient-based visual explainers (Grad-CAM variants) and image classification benchmarks. Because gradients can suffer from saturation, future research should integrate perturbation-based methods (e.g., SHAP) or propagation techniques (e.g., LRP). Additionally, investigating these dynamics in dense prediction tasks, such as object detection, will further clarify how spatial reasoning transfers during compression.

**Conclusion**

Deploying compact but opaque models in safety-critical domains poses significant risks. This study introduces an interpretability-aware evaluation framework for KD and shows that how much visual explainability a compressed model retains depends strongly on how the distillation objective is configured. Macro-level object focus can be preserved, and sometimes improved, when the soft-label weight is kept low. Fine-grained attribution proved harder to retain: it was compromised both by rigid convolutional bottlenecks and by excessive soft-label weighting. By prioritizing hard targets, minimizing temperature scaling, and leveraging the representational flexibility of Vision Transformers, developers can effectively preserve the spatial reasoning that renders neural networks transparent and trustworthy.

**TABLES**

Table 1: Cross-architecture distillation results with relative percentage changes in parentheses. Bold - Best score per student family.

| Teacher | Student | Acc1 | Acc5 | GC Mass | GGC Mass | GC Rank | GGC Rank |
|---|---|---|---|---|---|---|---|
| - | resnet34 | 73.31 | ***91.42*** | 0.063 | ***0.2284*** | 0.049 | ***0.234*** |
| resnet152 | resnet34 | 73.27 (-0.1%) | 91.39 (-0.0%) | ***0.083*** (+31.7%) | 0.1851 (-19.0%) | 0.094 (+91.8%) | 0.173 (-26.1%) |
| efficientnet_b7 | resnet34 | ***73.47*** (+0.2%) | 91.40 (-0.0%) | ***0.083*** (+31.7%) | 0.1870 (-18.1%) | ***0.097*** (+98.0%) | 0.180 (-23.1%) |
| vit_l_16 | resnet34 | 73.15<br>(-0.2%) | 91.27<br>(-0.2%) | 0.082<br>(+30.2%) | 0.1830<br>(-19.9%) | 0.089<br>(+81.6%) | 0.172<br>(-26.5%) |
| - | efficientnet_b0 | ***77.69*** | ***93.53*** | ***0.086*** | 0.0450 | 0.076 | 0.059 |
| resnet152 | efficientnet_b0 | 77.25<br>(-0.6%) | 93.38<br>(-0.2%) | 0.070<br>(-18.6%) | ***0.0600***<br>(+33.3%) | ***0.084***<br>(+10.5%) | ***0.072***<br>(+22.0%) |
| efficientnet_b7 | efficientnet_b0 | 77.18<br>(-0.7%) | 93.36<br>(-0.2%) | 0.077<br>(-10.5%) | 0.0590<br>(+31.1%) | 0.083<br>(+9.2%) | 0.071<br>(+20.3%) |
| vit_l_16 | efficientnet_b0 | 77.10<br>(-0.8%) | 93.20<br>(-0.4%) | 0.073<br>(-15.1%) | 0.06<br>(+33.3%) | 0.082<br>(+7.9%) | 0.071<br>(+20.3%) |
| - | vit_s_16 | 74.63 | 92.67 | 0.052 | 0.0690 | 0.072 | 0.084 |
| resnet152 | vit_s_16 | 80.82<br>(+8.3%) | ***95.92***<br>(+3.5%) | 0.079<br>(+51.9%) | ***0.0870***<br>(+26.1%) | 0.091<br>(+26.4%) | ***0.087***<br>(+3.6%) |
| efficientnet_b7 | vit_s_16 | ***80.83***<br>(+8.3%) | 95.85<br>(+3.4%) | 0.073<br>(+40.4%) | 0.0840<br>(+21.7%) | 0.078<br>(+8.3%) | 0.083<br>(-1.2%) |
| vit_l_16 | vit_s_16 | 79.50<br>(+6.5%) | 94.90<br>(+2.4%) | ***0.100***<br>(+92.3%) | 0.0320<br>(-53.6%) | ***0.150***<br>(+108.3%) | 0.037<br>(-56.0%) |

| - | mobilev4 small | ***73.73*** | 91.38 | 0.069 | ***0.1589*** | 0.069 | ***0.140*** |
|---|---|---|---|---|---|---|---|
| resnet152 | mobilev4 small | 73.49 (-0.3%) | ***91.40*** (+0.0%) | 0.070 (+1.4%) | 0.1250 (-21.3%) | 0.076 (+10.1%) | 0.115 (-17.9%) |
| efficientnet_b7 | mobilev4 small | 73.39 (-0.5%) | 91.35 (-0.0%) | ***0.075*** (+8.7%) | 0.1230 (-22.6%) | ***0.076*** (+10.1%) | 0.119 (-15.0%) |
| vit_l_16 | mobilev4 small | 72.48 (-1.7%) | 90.79 (-0.6%) | 0.072 (+4.3%) | 0.1230 (-22.6%) | 0.075 (+8.7%) | 0.117 (-16.4%) |
| vit_l_16 | - | 79.66 | 94.63 | 0.1 | 0.032 | 0.151 | 0.038 |
| resnet152 | - | 78.31 | 94.05 | 0.089 | 0.339 | 0.088 | 0.252 |
| efficientnet_b7 | - | 84.12 | 96.91 | 0.075 | 0.045 | 0.090 | 0.050 |

Table 2: Impact of temperature (T) and soft-label loss weight (α) on interpretability with Teacher and Student

| **Config** | **T** | α | **Acc1 (%)** | **GC Mass** | **GGC Mass** | **GC Rank** | **GGC Rank** |
|---|---|---|---|---|---|---|---|
| Teacher | - | - | 82.284 | - | - | - | - |

| (ResNet152) | | | | | | | |
|---|---|---|---|---|---|---|---|
| Student<br>Baseline | - | - | 73.314 | 0.0915 | 0.2027 | *0.1002* | 0.2041 |
| Config 1 | 1 | 0.5 | 73.953<br>(+0.9%) | ***0.0969***<br>(+5.9%) | 0.1853<br>(-8.6%) | 0.0993<br>(-0.9%) | ***0.2149***<br>(+5.3%) |
| Config 2 | 4 | 0.5 | 73.346<br>(+0.0%) | 0.0822<br>(-10.2%) | 0.1808<br>(-10.8%) | 0.0933<br>(-6.9%) | 0.1717<br>(-15.9%) |
| Config 3 | 10 | 0.5 | 73.369<br>(+0.1%) | 0.0839<br>(-8.3%) | 0.1854<br>(-8.5%) | 0.0955<br>(-4.7%) | 0.1813<br>(-11.2%) |
| Config 4 | 4 | 0.1 | ***74.023***<br>(+1,0%) | *0.0952*<br>(+4.0%) | ***0.1864***<br>(-8.0%) | ***0.1005***<br>(+0.3%) | 0.2134<br>(+4.6%) |
| Config 5 | 4 | 0.9 | 71.631<br>(-2.3%) | 0.0767<br>(-16.2%) | 0.1612<br>(-20.5%) | 0.0733<br>(-26.8%) | 0.1586<br>(-22.3%) |

**FIGURES**

Figure 1: General Architecture of Knowledge Distillation

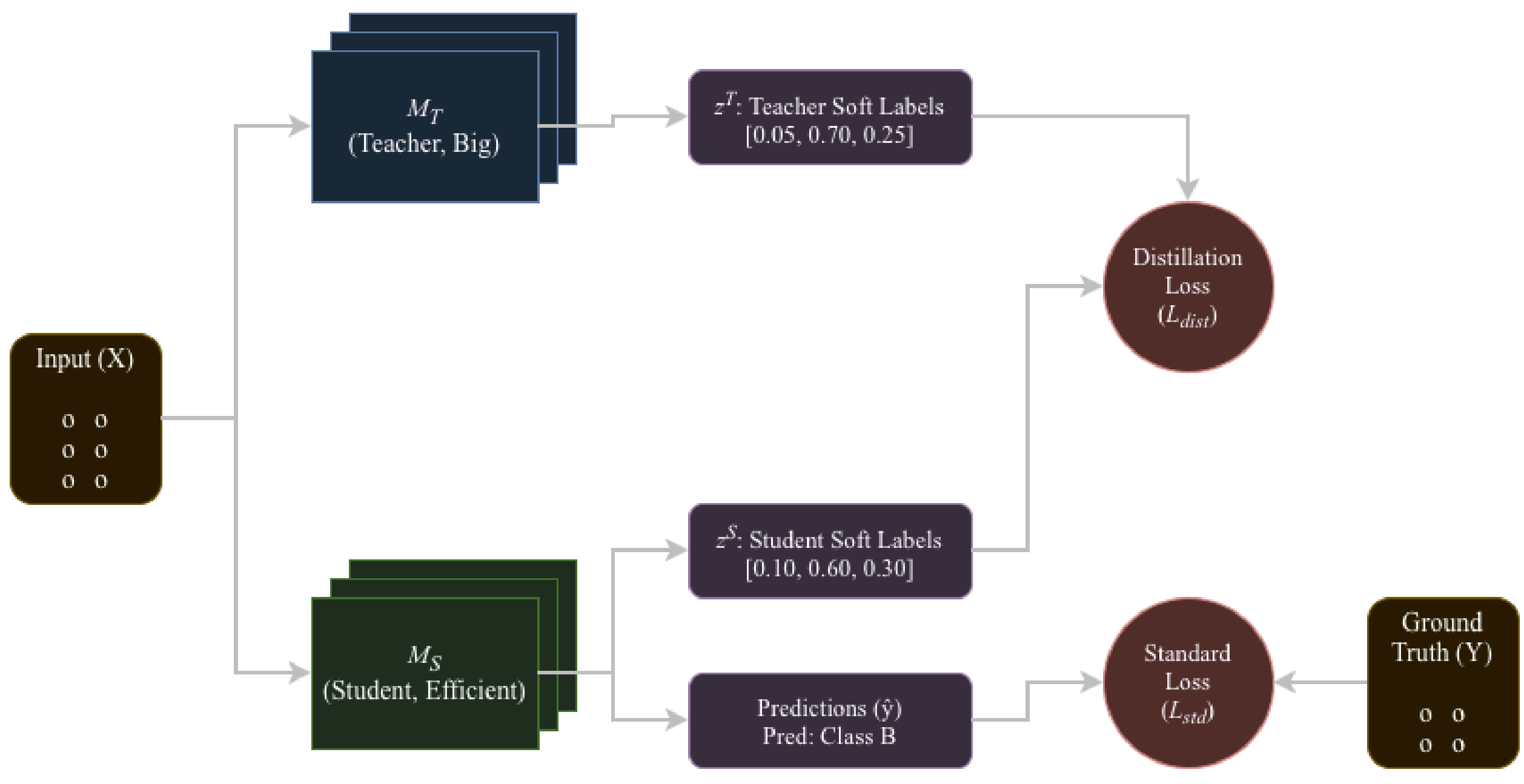

Figure 2: Baseline ResNet-34 vs. ResNet-152 to ResNet-34

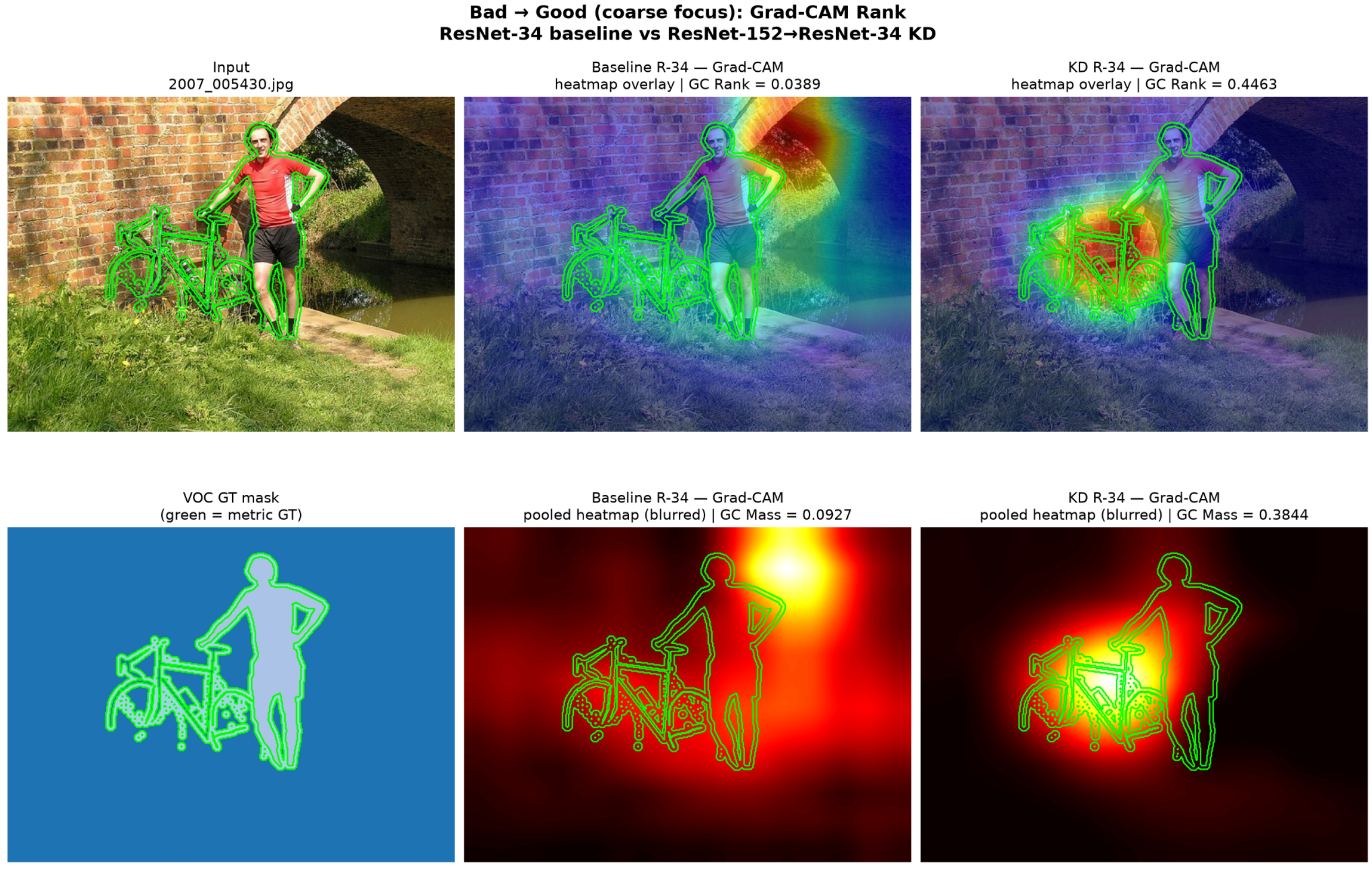

Figure 3: Coarse localization improves after distillation

**Good → Bad (fine-grained loss): Guided Grad-CAM Mass**
**ResNet-34 baseline vs ResNet-152→ResNet-34 KD**

Input
2008_002471.jpg

Baseline R-34 — Guided Grad-CAM
top-50% mass region | GGC Mass = 0.6636

KD R-34 — Guided Grad-CAM
top-50% mass region | GGC Mass = 0.0135

VOC GT mask
(green = metric GT)

Baseline R-34 — Guided Grad-CAM
pooled heatmap (blurred) | GGC Rank = 0.4472

KD R-34 — Guided Grad-CAM
pooled heatmap (blurred) | GGC Rank = 0.0552

Figure 4: Fine-grained attribution degrades after distillation

**Bad → Good (dual gain): GC Mass & GGC Mass**
**ViT-S/16 baseline vs ResNet-152→ViT-S/16 KD**

Input
2007_007930.jpg

Baseline ViT-S/16 — Grad-CAM
heatmap overlay | GC Mass = 0.0714

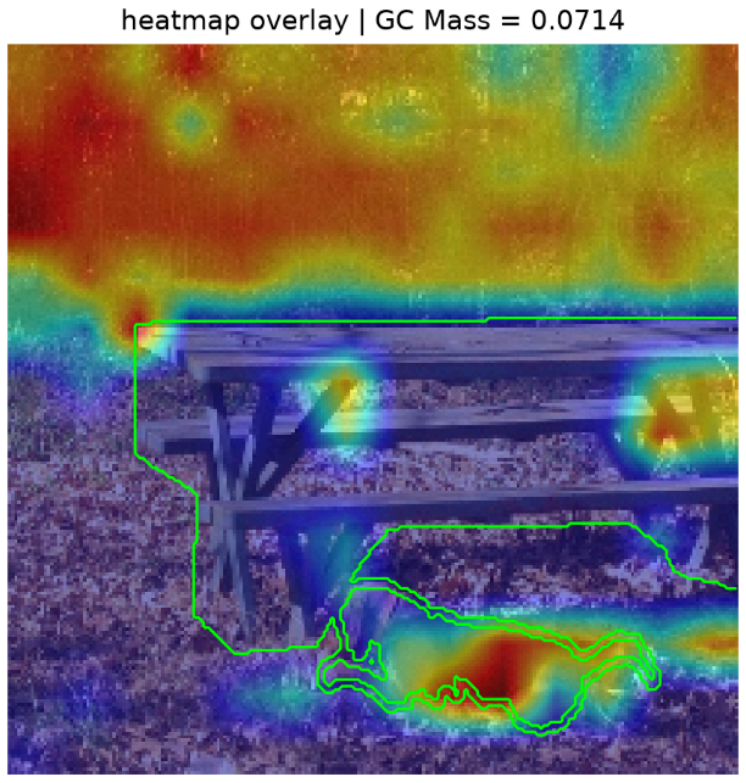

KD ViT-S/16 — Grad-CAM
heatmap overlay | GC Mass = 0.7715

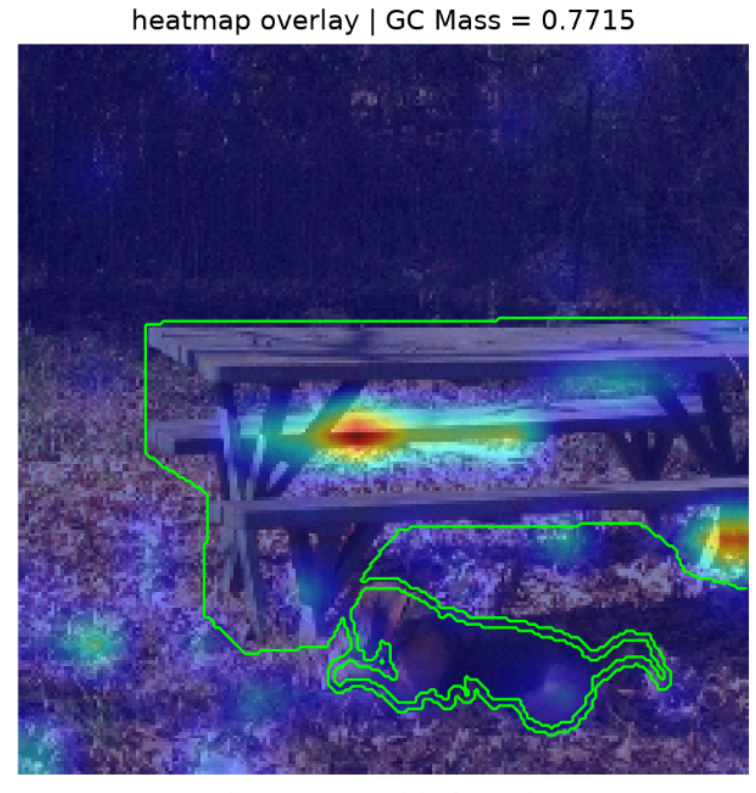

VOC GT mask
(green = metric GT)

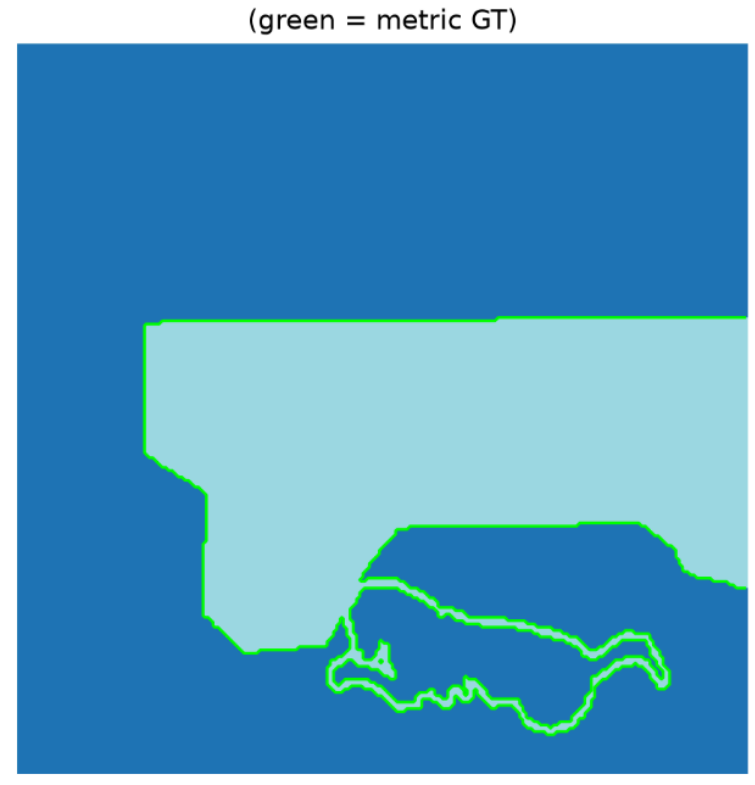

Baseline ViT-S/16 — Guided Grad-CAM
pooled heatmap (blurred) | GGC Mass = 0.1517

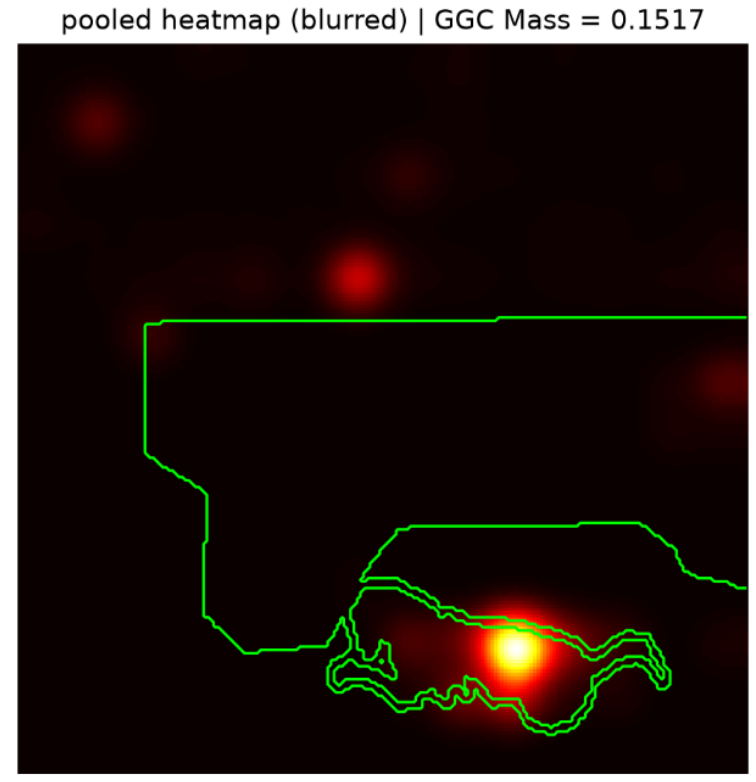

KD ViT-S/16 — Guided Grad-CAM
pooled heatmap (blurred) | GGC Mass = 0.4537

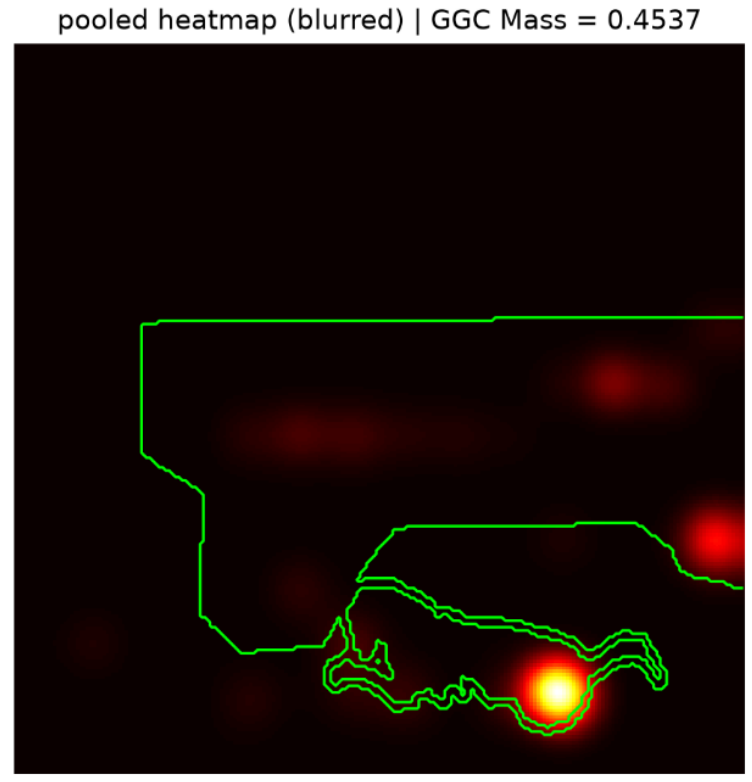